\documentclass{article}

 \usepackage[preprint]{neurips_2026}

\usepackage[utf8]{inputenc} 
\usepackage[T1]{fontenc}    
\usepackage{hyperref}       
\usepackage{url}            
\usepackage{booktabs}       
\usepackage{amsfonts}       
\usepackage{nicefrac}       
\usepackage{microtype}      
\usepackage{xcolor}         
\usepackage{amsmath}
\usepackage{graphicx}
\usepackage{multirow}
\usepackage{enumitem}
\usepackage{wrapfig}  
\usepackage{algorithm}
\usepackage{algorithmic}
\usepackage{xspace}

\title{Multi-turn RL with Structural and Performance Aware Rewards for CUDA Kernel Generation}

\author{%
  Quazi Ishtiaque Mahmud\textsuperscript{1}, Nesreen K. Ahmed\textsuperscript{2}, Ali Jannesari\textsuperscript{1}\\
  \textsuperscript{1} \textit{Iowa State University, Ames, IA, USA}\\
  \{mahmud, jannesar\}@iastate.edu \\
  \textsuperscript{2} \textit{Cisco AI Research, San Jose, CA, USA}\\
  nesahmed@cisco.com
}

\begin{document}

\newcommand{\ourtool}{\textsc{CUDAPerf}}

\maketitle

\begin{abstract}
  Reinforcement Learning with Verifiable Rewards (RLVR) has emerged as a powerful technique to enhance the reasoning capacity of LLMs for optimized code generation. However, existing RLVR approaches primarily rely on outcome-based signals such as correctness and speedup, overlooking performance-critical structural properties of programs that are essential for generating optimized code. In this work, we propose \ourtool, a reflective RL framework that incorporates both verifiable execution rewards and structural code-aware rewards derived from parallelization features (e.g., memory coalescing, occupancy, Arithmatic Intensity, and synchronization patterns). \ourtool \xspace{} operates in two stages: (1) an offline pairwise ranking module that learns to distinguish strong and weak program candidates via contrastive comparisons, and (2) an online RL training phase that jointly optimizes for correctness, performance, and structural efficiency through a unified reward signal. To further enhance learning, \ourtool \xspace{} utilizes iterative refinement using execution feedback enabling progressive improvement of generated candidates. We also introduce a dataset comprising 2.9k C $\rightarrow$ CUDA and 1k PyTorch $\rightarrow$ CUDA programs, each paired with diverse input configurations and multiple CUDA implementations encompassing diverse optimization strategies. \ourtool \xspace{} is evaluated across multiple benchmarks comprising both C $\rightarrow$ CUDA and PyTorch $\rightarrow$ CUDA transformations. Empirical findings suggest that \ourtool \xspace{} significantly outperforms strong baselines, including Qwen-3-32B (for C $\rightarrow$ CUDA) and CUDA Agent (for PyTorch $\rightarrow$ CUDA) by achieving up to 5X \& 3.32X improvements in speedup, and 17\% \& 7\% improvements in correctness, respectively.

\end{abstract}

\section{Introduction}

Automatic generation of high-performance parallel programs remains a fundamental challenge in high-performance computing (HPC). While modern large language models (LLMs) have demonstrated strong capabilities in code synthesis, they often struggle to produce correct and performant GPU programs, particularly for low-level parallel programming models such as CUDA. The difficulty arises from the need to jointly reason about functional correctness and complex parallel execution patterns, which are typically guided by expert knowledge and iterative profiling.

Recent advances in reinforcement learning with verifiable rewards (RLVR) have shown promise in improving reasoning tasks, including code generation and optimization. Recently, a large volume of works use RLVR and Agentic AI workflows to generate efficient GPU kernels \citep{baronio2025kevin, li2025cuda, du2025akg, liao2025kernelevolve, lange2025towards}. However, existing approaches largely rely on binary or sparse signals such as compilation success or correctness, which are insufficient for guiding models toward performance-optimal parallel implementations. In contrast, expert GPU programmers rely not only on execution feedback but also on structural properties of code, such as memory access patterns, thread mapping strategies, and synchronization behavior, which strongly influence performance \citep{nvidia_cuda_guide}. 

In this work, we propose a novel reinforcement learning framework \ourtool \xspace{} for generating high-performance CUDA programs from sequential code. Our approach integrates execution harness–based verifiable rewards with structure-aware signals derived from fine-grained program features, enabling the model to learn both what works (via execution feedback) and why it works (via structural patterns). To further improve learning efficiency, we introduce an offline pairwise ranking mechanism that learns to distinguish strong and weak candidates based on performance and leverages this signal to guide policy optimization.

We evaluate \ourtool \xspace{} on program translation tasks spanning C $\rightarrow$ CUDA and PyTorch $\rightarrow$ CUDA, demonstrating consistent improvements in both correctness and execution efficiency over strong baselines. Our results highlight the importance of combining verifiable execution feedback with structure-aware inductive biases for scalable and effective program optimization, bridging the gap between neural code generation and expert-level performance engineering. In summary, the paper makes the following contributions:
\vspace{-8pt}
\begin{itemize}
    \item We propose \textbf{\ourtool, a reinforcement learning framework} for generating high-performance CUDA programs that integrates execution-based verifiable rewards with structure-aware code features. \ourtool \xspace{} naturally extends to a multi-task setting, supporting both C $\rightarrow$ CUDA and PyTorch $\rightarrow$ CUDA translation. Empirical results demonstrate that \ourtool \xspace{} achieves up to \textbf{5$\times$} and \textbf{6$\times$} speedups over strong baselines such as Qwen-3-32B and CUDA-Agent on these tasks, respectively. 
    \item We propose a unified learning strategy that combines an \textbf{offline pairwise ranking mechanism} with \textbf{online RL optimization} to construct a composite reward signal. The offline stage learns to distinguish strong and weak program candidates via contrastive comparisons, while the online phase jointly optimizes correctness, performance, and structural efficiency by integrating execution feedback with structure-aware rewards. Empirical results show that combining structural and verifiable rewards yields more optimized CUDA programs.

    \item We curate a dataset comprising \textbf{2.9k C $\rightarrow$ CUDA} and \textbf{1k PyTorch $\rightarrow$ CUDA} programs, each paired with diverse input configurations and multiple CUDA implementations encompassing diverse optimization strategies to enable robust evaluation of both correctness and performance.
\end{itemize}

\section{Methodology}

\begin{figure}
    \centering
    \includegraphics[width=\textwidth]{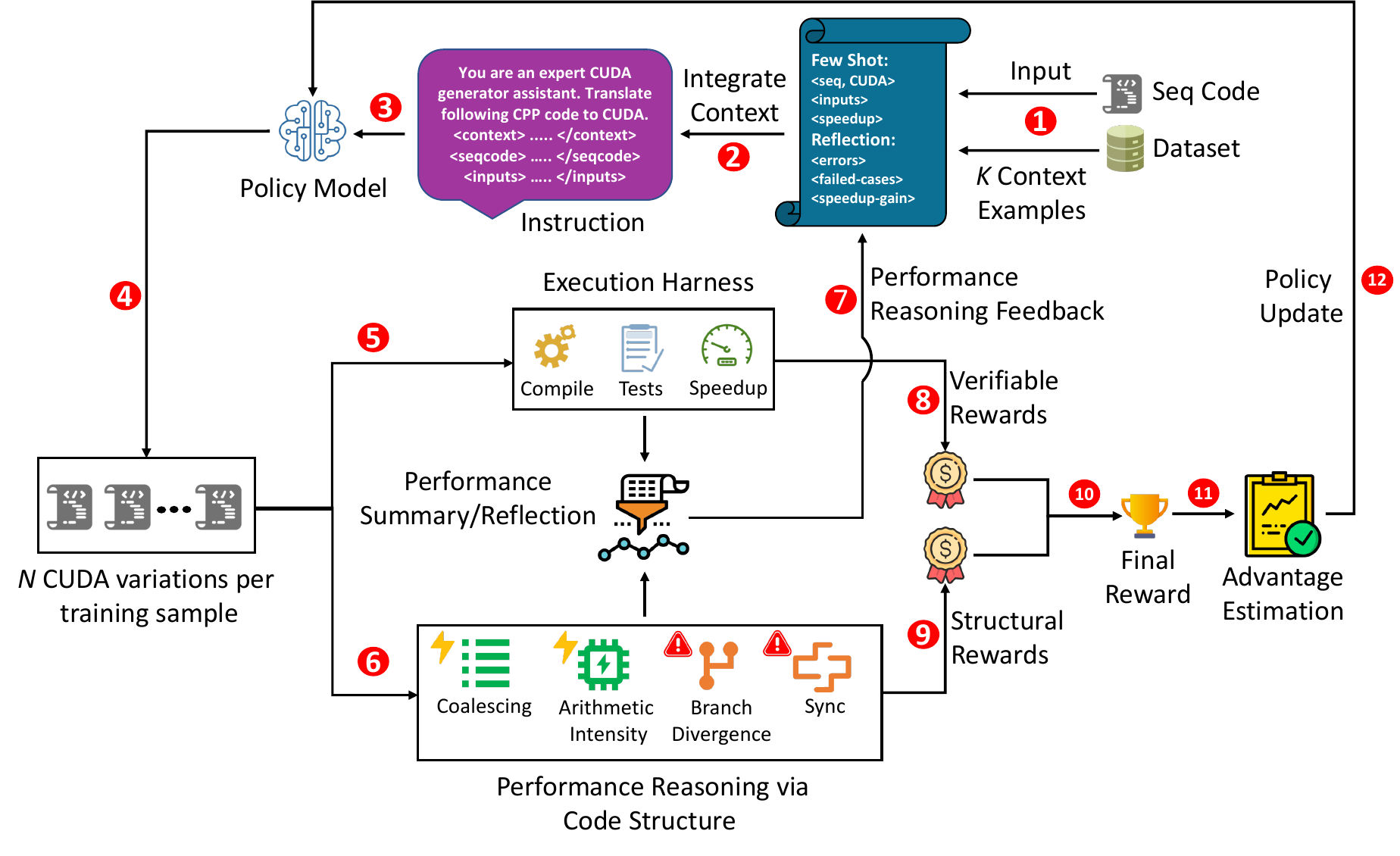}
    \caption{\textbf{\ourtool \xspace{} workflow.} The framework generates multiple CUDA candidates from sequential programs, evaluates them using execution-based verifiable rewards and structure-aware performance features, and iteratively refines the policy using reflection-driven feedback and reinforcement learning updates.}
    \label{figs:workflow}
\end{figure}

\ourtool \xspace{} works in two stages. Firstly, an offline pairwise ranking module is trained that learns to distinguish between strong and weak CUDA candidates via a scoring mechanism that favors optimized CUDA kernels. Then this ranking module is combined with execution based signals in online RL training phase to guide policy optimization. Figure \ref{figs:workflow} shows the workflow of \ourtool.



\subsection{Offline Training}

\subsubsubsection{\textbf{Structural Reward}}

One of the key contribution of this work is the integration of performance-relevant structural features in the reward. Drawing on the Nvidia CUDA guide \citep{nvidia_cuda_guide} and prior literature \citep{williams2009roofline, volkov2010better, hong2009analytical, grewe2014mapping, fung2007dynamic, cui2012accurate, pai2013improving} on performance modeling and optimization on GPU a set of core parallel code features that significantly influence program performance are identified that includes memory coalescing, arithmetic intensity \& operation count, occupancy (threads per block), control divergence, synchronization, atomic operations and data transfer overhead. While prior works in systems and architecture studied these factors, existing LLM-based code generators and RL-driven approaches largely ignore such structural signals and rely primarily on execution-based feedback \citep{liao2025kernelevolve, du2025akg, lange2025towards, baronio2025kevin, dai2026cuda, li2025cuda}.

To incorporate the mentioned CUDA structural properties into the reward function in RL, an offline lightweight ranker is trained to map extracted code features to a scalar performance score. More details of the feature extraction techniques is described in Appendix \ref{app:feature-extraction}. The feature vector looks like following,
\[
\phi(y) =
[\texttt{coal}, \texttt{ai}, \texttt{occ}, \texttt{div}, \texttt{xfer},
\texttt{atomics}, \texttt{syncthreads}, \texttt{kernels}, \texttt{tpb},
\texttt{gmem\_access}, \texttt{ops}]
\]


Then feature normalization is applied $\hat{\phi}_j(y) = \frac{\phi_j(y) - \mu_j}{\sigma_j}$; where $\mu_j$ and $\sigma_j$ denote the mean and standard deviation of feature $j$ over the training split. Each training task contain multiple CUDA candidates. We sort the CUDA candidate programs for each task by measured speedup. The speedup is measured as the ratio of sequential and CUDA runtime. The bottom quartile is treated as the weak candidate set, while the top quartile is treated as the strong candidate set. We then sample pairwise preferences $ (y^+, y^-) $ where $y^+$ is sampled from the top-performing quartile and $y^-$ is sampled from the bottom-performing quartile, subject to $ \mathrm{speedup}(y^+) > \mathrm{speedup}(y^-) $. 


We train an MLP ranker $s_\psi(\hat{\phi}(y))$ that maps the normalized feature vector of a candidate program to a scalar structural score. The model consists of fully connected layers with ReLU activations and dropout, followed by a final linear layer producing a single score. The ranker is trained to assign higher scores to faster candidates within each pair.

Given a pair $(y^+, y^-)$, the ranker is optimized using the logistic pairwise ranking loss:
\[
\mathcal{L}_{\mathrm{rank}}(\psi)
=
\mathbb{E}_{(y^+,y^-)}
\left[
\log\left(1 + \exp\left(-(s_\psi(\hat{\phi}(y^+)) -
s_\psi(\hat{\phi}(y^-)))\right)\right)
\right].
\]
This objective encourages the score difference $ s_\psi(\hat{\phi}(y^+)) - s_\psi(\hat{\phi}(y^-)) $ to be positive whenever $y^+$ has higher measured speedup than $y^-$. In practice, this corresponds to applying the softplus loss to the negative score difference. The model is trained with AdamW, mini-batch optimization, weight decay, dropout regularization, and gradient clipping. After training, the ranker provides a learned structural reward:
\begin{equation}
    \label{eq:reward-struct}
    R_{\mathrm{str}}(y) = s_\psi(\hat{\phi}(y))
\end{equation}

This score in Eq. \ref{eq:reward-struct} is considered as \textbf{Structural Reward} during online reinforcement learning.

\subsection{Online RL Training}

\subsubsection{Problem Setup}

We formalize the task as learning a conditional policy $\pi_{\theta}(y \mid x)$, where x represents a sequential input program and y represents a candidate CUDA implementation where the policy is parametrized by $\theta$. For each input x the policy generates $g$ candidates $Y(x) = {y_1, y_2, ..., y_g}$. This enables group-relative evaluation. The objective is to learn an optimal policy $\pi_{\theta}^{*}$ (Eq. \ref{eq:eq-main-policy}), that maximizes the expected reward over generated programs, where the reward integrates both verifiable execution feedback $R_{\text{ver}}(y, x)$ and structure-aware performance signals $R_{\text{str}}(y, x)$.

\begin{equation}
    \pi_{\theta}^{*} = \arg\max_{\pi_{\theta}} \; \mathbb{E}_{y \sim \pi_{\theta}(\cdot \mid x)} \left[ R_{\text{ver}}(y, x) +  R_{\text{str}}(y, x) \right]
\label{eq:eq-main-policy}
\end{equation}

Here $R_{\mathrm{ver}}$ captures correctness and runtime efficiency while $R_{\mathrm{str}}$ provides a learned prior over performance-relevant CUDA code structure. These code structural features introduce structured inductive bias that complements execution-based feedback with code structure-based signals. This unified formulation of composite reward enables the model to reason not only from observed performance outcomes (from execution harness) but also from interpretable code-level properties, thereby facilitating a transition from purely black-box optimization to more principled, white-box reasoning over program performance.


\subsubsubsection{\textbf{Verifiable Reward}}

The verifiable reward is measured by passing each generated CUDA response through an execution harness that evaluates using compilation correctness, test case pass ratio and speedup gain. The verifiable reward consists of following components:



\textbf{Compilation Penalty.} A fixed hard penalty is applied if the candidate CUDA fails to compile.
\begin{equation*}
R_{\text{comp}}(y) =
\begin{cases}
-\alpha_{\text{comp}}, & \text{if compilation fails} \\
0, & \text{otherwise}
\end{cases}
\end{equation*}

\textbf{Correctness Penalty.} Incorrect outputs are penalized proportionally to the pass rate where pass rate is computed as $p(y)=\frac{\# \ passed \ tests}{\# \ total \ test \ cases}$.
\begin{equation*}
R_{\text{corr}}(y) = -\alpha_{\text{wrong}} \cdot (1 - p(y))
\end{equation*}

\textbf{Performance Reward.} For correct programs meaning programs that compile and also pass all test cases, we define the speedup as
$S = \frac{T_{\text{CPU}}(y)}{T_{\text{GPU}}(y)}$. Here $T_{\text{CPU}}(y)$ and $T_{\text{CPU}}(y)$ are the sequential program and CUDA program median runtime across 5 executions. Then the performance reward is computed as:
\begin{equation*}
R_{\text{speed}}(y) = \alpha_{\text{speed}} \cdot 
\text{clip}\big(\log_2(S), \, s_{\min}, \, s_{\max}\big)
\end{equation*}
The log-scaling and clipping is performed to transform the speedup reward to a bounded range which stabilizes training and prevent extreme performance outliers from becoming dominating entities in the reward signal.

\textbf{Stability Penalty.} To discourage unstable kernels, high runtime variance is penalized using the coefficient of variation $\mathrm{cv}(y)$:
\begin{equation*}
R_{\text{cv}}(y) = -\alpha_{\text{cv}} \cdot \mathrm{cv}(y),
\end{equation*}

The final \textbf{Verifiable Reward} is given by:
\begin{equation*}
R_{\text{ver}}(y, x)  =
\begin{cases}
R_{\text{comp}}(y), & \text{if compilation fails} \\
R_{\text{corr}}(y), & \text{if } p(y) < 1 \\
R_{\text{corr}}(y) + R_{\text{speed}}(y) + R_{\text{cv}}(y), & \text{if } p(y) = 1 \\
\end{cases}
\end{equation*}

We empirically set the values $\alpha_{\text{comp}} = 3.0$, $\alpha_{\text{wrong}} = 2.0$, $\alpha_{\text{speed}} = 1.0$ and $\alpha_{\text{cv}} = 0.5$. Also, to bound the speedup reward to moderate slowdowns and upto 16X speedup we set the $s_{\min} = -0.5$ and $s_{\max} = 4.0$.

\section{Experimental Results}

\vspace{-5pt}

\textbf{Base Models.} For the RL training the policy model is based on Qwen-3-32B \citep{qwen3technicalreport}. We finetune using QLoRA (4 bit) \citep{dettmers2023qlora}. We train on two A100 40 GB GPUs. The CUDA executions are pinned to a single GPU to avoid interference between training and benchmarking. We focus on two problems: i) C to CUDA and ii) PyTorch to CUDA transformation. Our training recipe remains almost identical for both of the problems.

\textbf{Execution Harness.} The execution harness consists of three components: compilation check module, test case check module and speedup calculation module. The compilation is checked to ensure that the generated CUDA candidates do not have any syntactic errors. For each sample in our dataset we have at least five test cases and the test case check module runs each candidate CUDA program through these test cases and checks how many of these test cases are passed. Finally, the speedup calculation module runs each candidate CUDA five times, takes the average runtime, calculates the speedup compared to the sequential program. 

\textbf{Training offline ranker.} Offline training is done only once to teach the ranker model to identify preferable CUDA kernels. We use existing dataset like BabelTower \citep{wen2022babeltower}, CUDA Agent \citep{dai2026cuda} to some extent which has sequential and their corresponding CUDA implementations along with test cases. But for training our offline ranker, we need to have a dataset where for each sequential program, there are multiple valid CUDA kernels such that the ranker can learn to identify which kernels are preferable. So, to increase diversity and meet our criterial we also used the CodeNet \citep{puri2021codenet} dataset that contains programs from AIZU \citep{aizu_online_judge} and AtCoder \citep{atcoder_online_judge} online judges, which includes thousands of programs along with test cases. Then, we employ LLM (Qwen-3-32B \citep{qwen3technicalreport}) to generate at least 10 different CUDA variations for each sequential program. We also employed LLM to generate randomized test inputs following the approach of \citep{baronio2025kevin} where necessary. We collected around 50k samples however we use a rigorous filtering where we only considered those samples where we have at least 5 CUDA candidates that successfully compile, pass all the test cases, and at least one variant has $\geq$1.5X speedup compared to the sequential program. This results in around 2903 C and 1013 PyTorch programs, where for each sequential (C/PyTorch) program, there are at least 5 valid CUDA candidates. 

For training the offline ranker we only need the CUDA variants for each sequential program as the goal is to increase ranking score for the CUDA kernels that are most performant. We employ a lightweight multi-layer perceptron (MLP) ranker to learn a scalar structural score for each CUDA program candidate before starting RL training. The network consists of fully connected layers with ReLU activations and dropout regularization. We use hidden dimensions of \(64\) and \(32\) with a dropout probability of \(0.1\). The model is optimized using AdamW with learning rate \(3 \times 10^{-4}\), weight decay \(10^{-4}\), batch size \(64\), and gradient clipping threshold \(1.0\). Training is performed for 50 epochs as we observe loss saturates after that. More details are provided in the Appendix \ref{app:ranker-arch}.

\textbf{Multi-turn RL Training.} The RL training is done separately for C$\rightarrow$CUDA and PyTorch$\rightarrow$CUDA. For the RL training we start from the base Qwen-3-32B \citep{qwen3technicalreport} as the policy model. For each training sample which is a sequential program we generate 16 candidate CUDA programs. Each candidate is passed through the execution harness and we collect the compilation status, test case pass ratio and also, the observed speedup compared to the sequential program. This provides us with the verifiable reward $R_{\text{ver}}$. Then for each candidate CUDA program the code structural properties are also extracted and we generate a structural reward score $R_{\text{str}}$ using the offline trained ranker. Both these rewards are combined to obtain the composite reward. Using these composite reward the policy model learns to identify the optimal CUDA candidates for each sequential program. We compute the GRPO loss according to Eq. \ref{eq:grpo-loss} directly adapted from \citep{guo2025deepseek} which updates the policy by maximizing the GRPO objective. 
\begin{equation*}
\mathcal{J}_{\text{GRPO}}(\theta) =
\mathbb{E}_{q \sim P(Q), \{o_i\}_{i=1}^{G} \sim \pi_{\theta_{\text{old}}}(\cdot \mid q)}
\end{equation*}
\begin{equation}
\begin{aligned}
\quad \left[
\frac{1}{G} \sum_{i=1}^{G} \frac{1}{|o_i|}
\sum_{t=1}^{|o_i|}
\min \left(
\frac{\pi_\theta(o_{i,t} \mid q, o_{i,<t})}
     {\pi_{\theta_{\text{old}}}(o_{i,t} \mid q, o_{i,<t})}
\, \hat{A}_{i,t},
\;
\text{clip}\left(
\frac{\pi_\theta(o_{i,t} \mid q, o_{i,<t})}
     {\pi_{\theta_{\text{old}}}(o_{i,t} \mid q, o_{i,<t})},
1 - \epsilon,\,
1 + \epsilon
\right)
\hat{A}_{i,t}
\right)
\right] \\
\quad - \beta D_{\mathrm{KL}}\big(\pi_\theta \,\|\, \pi_{\text{ref}}\big)
\end{aligned}
\label{eq:grpo-loss}
\end{equation}

Here $\hat{A}_{i,t} = \frac{r_i - \mathrm{mean}(r)}{\mathrm{std}(r)}$ and $r_i$ represents the composite reward. Hence,  $r_i = R_{\text{ver}}(i) + R_{\text{str}}(i)$; which means the combination of verifiable rewards and structural rewards for $i^{th}$ CUDA candidate. Here $G$ represents each group and $o_i$ represents each candidate CUDA program. $\pi_{\theta_{\text{old}}}$ and $\pi_{\theta}$ represents the old and the new policy model, respectively. We set the clipping parameter $\epsilon = 0.2$ which checks the policy updates and helps to prevent instability during training. Also, we set $\beta D_{\mathrm{KL}} = 0$ following \cite{baronio2025kevin} so that the policy model can deviate freely from the base model. For each training sample, we normalize the reward across 16 generated candidates and then the GRPO loss is computed for the entire batch following the approach of \cite{baronio2025kevin}.

\textbf{Structured Feedback and Iterative Refinement.} Also, for each candidate we take upto 5-refinement turns as the policy model may not be able to generate correct CUDA programs in the first attempt. In each refinement turn, we provide feedback from both execution harness and code structural properties. These feedbacks are incorporated in the instructions for the policy model. If a program fails to compile the compiler error messages are extracted and truncated to provide actionable debugging signals. If the program compiles but fails certain test cases then the failed cases are included in the feedback such that the policy model can readjust based on the test cases. If a program passes all test cases then we include the observed speedup in the feedback. Finally, we also include the code structural feature values in the feedback. These feedbacks are formatted as structured text and appended to the prompt. Iterative refinement is performed by conditioning the model on both previous generation and its associated feedback. To prevent context explosion for multiple refinement turns we discard context from previous turns and instead use another LLM (Qwen-3-32B \citep{qwen3technicalreport}) to summarize the changes that are applied to the previous turn and include the summary in the prompt.

\textbf{Prompt Construction.} The prompts are constructed by combining structured task content with perforamnce-aware few-shot examplers. For each task the prompt contains the sequential code and input test cases and these are provided using structured delimiters: [SEQ\_CODE], [INPUTS]. For constructing the few-shot pool we randomly sample $D_k$ samples from the dataset. Then these samples are sorted according to the speedup. From these sorted dataset only top $N$ candidates are retained. Finally we randomly sample $k$ examples as few-shot context. The few-shot context is given within a [EXAMPLE] block in the prompt. Each selected sample is formatted using explicit delimiters: [SEQ\_CODE], [CUDA], [INPUTS], [SPEEDUP]. The details of the prompts used for the instructions are described in the Appendix. We empirically set the values to $k=5$, $N=20$ and $D_k=50$.

We train for 60 epochs as we found rewards usually reach a plateau after that. Also, we sample 16 candidates from policy model at each turn using nucleus sampling with $p=0.9$ and temperatures [0.2, 0.7] to enable balance between conservative and diverse generations. The maximum generation length is set to 1536 tokens which is sufficient for capturing complete CUDA programs. For tackling out-of-memory issues during training we compute masked log-probabilities over a truncated window of 1024 tokens.

\textbf{Metrics.} As the prime focus of this work is to optimize CUDA kernels we evaluate our approach using correctness also the performance of the generated CUDA programs. The correctness is determined through testing the generate CUDA candidates using the test cases. If a CUDA candidate passes all the test cases, only then do we consider it a correct candidate. The second metric performance is calculated by taking the ratio of sequential and CUDA runtimes. We report performance using the geometric mean of speedups across tasks in the evaluation set.

\begin{table}[]
\centering
\caption{Comparison of correctness (all test cases passed) and performance (geometric mean of speedup) across multiple benchmarks. \ourtool \xspace{} consistently outperforms strong baselines in both functional correctness and execution efficiency. All results are reported for pass@5.}
\label{tab:main_results}
\resizebox{\linewidth}{!}{%
\begin{tabular}{cccccc}
\toprule
\textbf{Dataset} & \textbf{Task} & \textbf{Model} & \begin{tabular}[c]{@{}c@{}}\textbf{Correctness} \\ (All test cases passed)\end{tabular} & \begin{tabular}[c]{@{}c@{}}\textbf{Performance} \\ (Geo. Mean of Speedup)\end{tabular} \\ 
\midrule

\multirow{6}{*}{\ourtool \xspace{} Dataset (Ours)}  
& \multirow{6}{*}{C $\rightarrow$ CUDA}
& OpenAI O4-mini  \citep{openai_api}     & 41\% & 1.3X \\ 
& & CodeRosetta \citep{tehranijamsaz2024coderosetta}  & 45\% & 1.5X \\ 
& & CodeLlama-34b-Instruct-hf \citep{roziere2023code}  & 53\% & 3.5X \\ 
& & Qwen2.5-Coder-32B \citep{hui2024qwen2}    & 61\% & 5.3X \\ 
& & QiMeng-MuPa \citep{ke2025qimeng}     & 63\% & 5.9X \\ 
& & Qwen-3-32B \citep{qwen3technicalreport}     & 72\% & 6.2X \\ 
& & \textbf{\ourtool \xspace{} (Ours)}        & \textbf{89\%} & \textbf{11.02X} \\ 
\midrule

\multirow{6}{*}{BabelTower} 
& \multirow{6}{*}{C $\rightarrow$ CUDA}
& OpenAI O4-mini  \citep{openai_api}     & 52\% & 1.9X \\ 
& & CodeLlama-34b-Instruct-hf \citep{roziere2023code} & 68\% & 3.1X \\ 
& & CodeRosetta \citep{tehranijamsaz2024coderosetta}  & 71\% & 3.2X \\ 
& & Qwen2.5-Coder-32B \citep{hui2024qwen2}    & 79\% & 4.4X \\ 
& & Qwen-3-32B \citep{qwen3technicalreport}     & 86\% & 4.92X \\ 
& & \textbf{QiMeng-MuPa} \citep{ke2025qimeng}     & \textbf{94\%} & 6.01X \\ 
& & \textbf{\ourtool \xspace{} (Ours)}        & \textbf{94\%} & \textbf{9.02X} \\ 
\midrule

\multirow{6}{*}{KernelBench}      
& \multirow{6}{*}{PyTorch $\rightarrow$ CUDA}
& OpenAI O4-mini  \citep{openai_api}   & 38\% & 0.78X \\ 
& & CodeLlama-34b-Instruct-hf \citep{roziere2023code} & 41\% & 0.25X \\ 
& & Qwen2.5-Coder-32B \citep{hui2024qwen2}    & 51\% & 0.5X \\ 
& & Qwen-3-32B  \citep{qwen3technicalreport}    & 56\% & 0.53X \\ 
& & Kevin \citep{baronio2025kevin}          & 82\% & 1.10X \\ 
& & Cuda Agent \citep{dai2026cuda}     & 86\% & 3.09X \\ 
& & \textbf{\ourtool \xspace{} (Ours)}         & \textbf{93\%} & \textbf{6.41X} \\ 

\bottomrule
\end{tabular}%
}

\end{table}

\subsection{C to CUDA translation}

\begin{wrapfigure}{r}{0.48\columnwidth}
\vspace{-53pt}
    \centering
    \includegraphics[width=\linewidth]{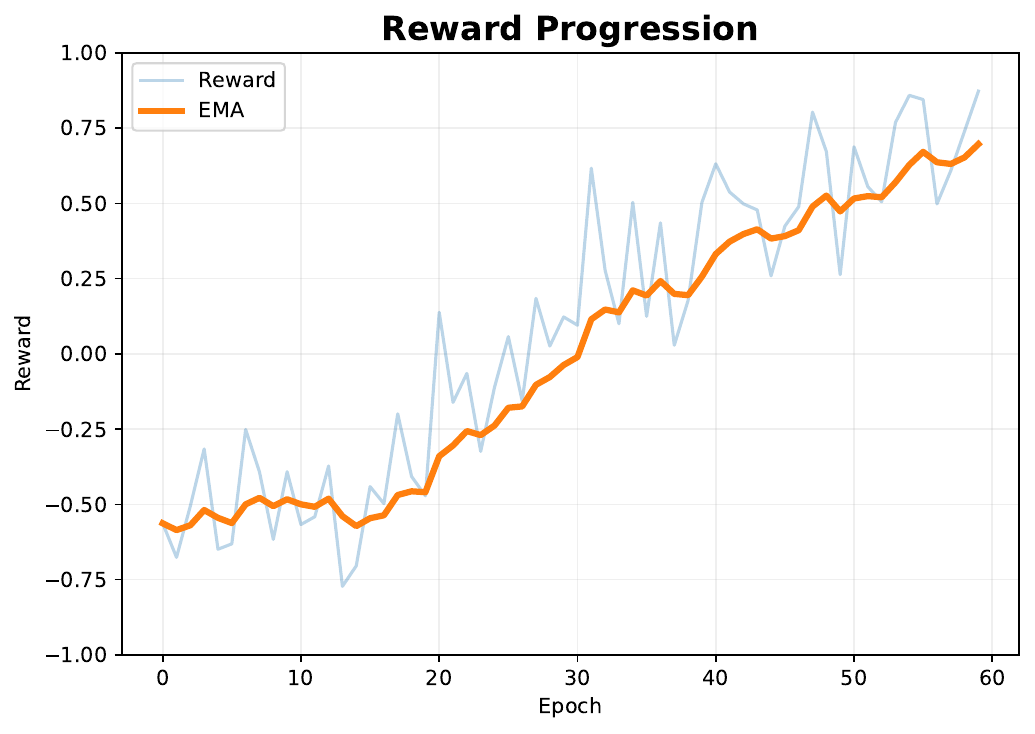}
    \vspace{-20pt}
    \caption{Reward Progression over epochs for C-to-CUDA translation. (\ourtool \xspace{} Dataset)}
    \label{figs:reward-cuda}
    \vspace{-20pt}
\end{wrapfigure}

\textbf{Dataset.} For RL Training on this task we use the curated dataset of 2903 C-to-CUDA programs. We keep around 10\% of the dataset (290 samples) for evaluation. In addition to that, we also evaluate on the BabelTower \citep{wen2022babeltower} evaluation set of 233 pairs of C and CUDA programs. 

\textbf{Baselines.} As baselines we compare with state-of-the-art open-source model Qwen-3-32B \citep{qwen3technicalreport}, Qwen2.5-Coder-32B \citep{hui2024qwen2}, CodeLlama-34b-Instruct-hf \citep{roziere2023code} and closed source model OpenAI O4-mini \citep{openai_api}. In addition to that, we also compare with task specific models state-of-the-art QiMeng-MuPa \citep{ke2025qimeng} and CodeRosetta \citep{tehranijamsaz2024coderosetta} which are specifically designed for C to CUDA translation.

\textbf{Results.} From Table \ref{tab:main_results} it can be observed that \ourtool \xspace{} achieves state-of-the-art results on both our and also the BabelTower dataset. On both datasets, \ourtool \xspace{} outperforms closed source OpenAI O4-mini \citep{openai_api} by achieving 48\% \& 42\% better results in terms of correctness. Also, on both datasets \ourtool \xspace{} achieves 9.72X \& 5.23X better speedup compared to OpenAI O4-mini \citep{openai_api}. Also, on both datasets \ourtool \xspace{} achieves 17\% \& 8\% better results in terms of correctness and 4.82X \& 4.1X better speedup than the Qwen-3-32B \citep{qwen3technicalreport} model. Also, on our dataset \ourtool \xspace{} outperforms current state-of-the-art model QiMeng-MuPa \citep{ke2025qimeng} by achieving 26\% better correctness and 5.12X better speedup. In BabelTower dataset \ourtool \xspace{} achieves comparable performance with Qimeng-MUPA \citep{ke2025qimeng} in terms of correctness. However, \ourtool \xspace{} achieves 3.01X better speedup compared to QiMeng-MuPa \citep{ke2025qimeng} which indicates that our RL based training along with verifiable and structural rewards helps to generate more efficient CUDA kernels. We provide more detailed results in Appendix \ref{app:results-pass}.

\subsection{PyTorch to CUDA translation}

\begin{wrapfigure}{r}{0.48\columnwidth}
\vspace{-50pt}
    \centering
    \includegraphics[width=\linewidth]{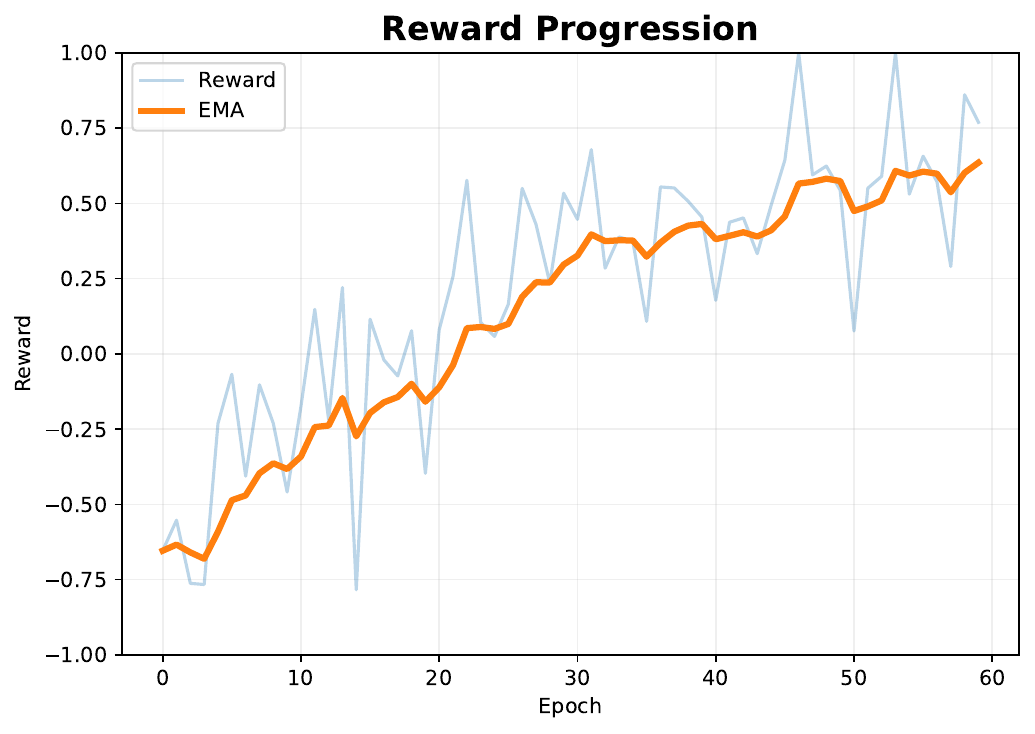}
    \vspace{-20pt}
    \caption{Reward Progression over epochs for PyTorch-to-CUDA translation. (KernelBench Dataset)}
    \label{figs:reward-pytorch}
    \vspace{-10pt}
\end{wrapfigure}

\textbf{Dataset.} We use the curated dataset which contains 1013 PyTorch-CUDA  programs for RL training on this task. For evaluation we use the KernelBench \cite{ouyang2025kernelbench} dataset which contains 250 PyTorch programs consisting of mostly ML workloads across three different levels. 


\textbf{Baselines.} We choose Qwen2.5-Coder-32B \citep{hui2024qwen2}, Qwen-3-32B \citep{qwen3technicalreport}, CodeLlama-34b-Instruct-hf \citep{roziere2023code} as open-source and OpenAI O4-mini \citep{openai_api} as closed source baselines. For task specific model we choose Cuda Agent \citep{dai2026cuda} and Kevin \citep{baronio2025kevin}, which are state-of-the-art models specifically designed for PyTorch to CUDA translation.

\textbf{Results.} Results from Table \ref{tab:main_results} show that \ourtool \xspace{} performs better than specilized models like CUDA Agent \citep{dai2026cuda} and Kevin \citep{baronio2025kevin}. \ourtool \xspace{} achieves 11\% and 7\% better results in terms of correctness than Kevin \citep{baronio2025kevin} and CUDA Agent \citep{dai2026cuda}, respectively. Also, it achieves 3.32X better speedup compared to the best performing model CUDAAgent. \ourtool \xspace{} outperforms OpenAI O4-mini \citep{openai_api} by achieving 55\% better correctness and a significant 5.63X speedup gain. \ourtool \xspace{} also achieves 37\% better correctness and 5.88X better speedup than the Qwen-3-32B \citep{qwen3technicalreport} model. We provide more detailed results in Appendix \ref{app:results-pass}.

\subsection{Ablation on Model Size} 

We experimented with different model sizes as starting point for our \ourtool \xspace{} training. The results are reported in Table \ref{tab:ablation-size}. Each time we use a different size for Qwen-3-32B \citep{qwen3technicalreport} model as our base policy model in RL. It can be observed from Table \ref{tab:ablation-size} that the increasing the model size also results in better correctness and performance results for \ourtool. This is also consistent with the findings of \citep{baronio2025kevin} paper that reported using a stronger prior helps to obtain better results. Hence, we use Qwen-3-32B \citep{qwen3technicalreport} as our base policy model for RL training.

\begin{table}[]
\centering
\caption{Comparing performance of \ourtool \xspace{} for different sizes of policy model. We choose 3 different Qwen-3 model \citep{qwen3technicalreport} sizes (8B, 14B and 32B). The base policy model that we use for RL training is given in square brackets. All results are reported for pass@5.}
\label{tab:ablation-size}
\resizebox{\linewidth}{!}{%
\begin{tabular}{cccccc}
\toprule
\textbf{Dataset} & \textbf{Task} & \textbf{Model} & \begin{tabular}[c]{@{}c@{}}\textbf{Correctness} \\ (All test cases passed)\end{tabular} & \begin{tabular}[c]{@{}c@{}}\textbf{Performance} \\ (Geo. Mean of Speedup)\end{tabular} \\ 
\midrule

\multirow{3}{*}{\ourtool \xspace{} Dataset (Ours)}  
& \multirow{3}{*}{C $\rightarrow$ CUDA}
&   \ourtool \xspace{} (Qwen-3-8B)     & 69\% & 6.9X \\ 
& & \ourtool \xspace{} (Qwen-3-14B)    & 77\% & 9.1X \\ 
& & \textbf{\ourtool \xspace{} (Qwen-3-32B)}    & \textbf{89\%} & \textbf{11.02X} \\  
\midrule

\multirow{3}{*}{BabelTower} 
& \multirow{3}{*}{C $\rightarrow$ CUDA}
&   \ourtool \xspace{} (Qwen-3-8B)     & 76\% & 4.3X \\ 
& & \ourtool \xspace{} (Qwen-3-14B)    & 81\% & 5.9X \\ 
& & \textbf{\ourtool \xspace{} (Qwen-3-32B)}    & \textbf{94\%} & \textbf{9.02X} \\  
\midrule

\multirow{3}{*}{KernelBench}      
& \multirow{3}{*}{PyTorch $\rightarrow$ CUDA}
&   \ourtool \xspace{} (Qwen-3-8B)     & 71\% & 3.4X \\ 
& & \ourtool \xspace{} (Qwen-3-14B)    & 83\% & 4.5X \\ 
& & \textbf{\ourtool \xspace{} (Qwen-3-32B)}    & \textbf{93\%} & \textbf{6.41X} \\  

\bottomrule
\end{tabular}%
}

\end{table}

\subsection{Ablation on Verifiable and Structural Rewards} 

We also perform experiments by only using one of the reward signals and masking off the other one. Table \ref{tab:ablation-feature} shows the results. It can be observed that both components contribute significantly in achieving better correctness and speedup results. Without structural rewards, the correctness drops 8\%, 7\% \& 9\%, and performance drops 1.72X, 2.92X \& 2.11X across the \ourtool, BabelTower and KernelBench dataset respectively. And, without verifiable rewards, the correctness drops 14\%, 13\% \& 17\%, and performance drops 4.12X, 4.42X \& 2.71X across the \ourtool, BabelTower and KernelBench dataset respectively. 

\begin{table}[h]
\centering
\caption{Analyzing effects on Verifiable and Structural rewards in \ourtool. We use the Qwen-3-32B \citep{qwen3technicalreport} model as the RL base policy model. All results are reported for pass@5.}
\label{tab:ablation-feature}
\resizebox{\linewidth}{!}{%
\begin{tabular}{cccccc}
\toprule
\textbf{Dataset} & \textbf{Task} & \textbf{Model} & \begin{tabular}[c]{@{}c@{}}\textbf{Correctness} \\ (All test cases passed)\end{tabular} & \begin{tabular}[c]{@{}c@{}}\textbf{Performance} \\ (Geo. Mean of Speedup)\end{tabular} \\ 
\midrule

\multirow{3}{*}{\ourtool \xspace{} Dataset (Ours)}  
& \multirow{3}{*}{C $\rightarrow$ CUDA}
&  \ourtool \xspace{} (with structural reward only)    & 75\% & 6.9X \\ 
& &  \ourtool \xspace{} (with verifiable reward only)     & 81\% & 9.3X \\ 
& & \textbf{\ourtool \xspace{}}    & \textbf{89\%} & \textbf{11.02X} \\  
\midrule

\multirow{3}{*}{BabelTower} 
& \multirow{3}{*}{C $\rightarrow$ CUDA}
&   \ourtool \xspace{} (with structural reward only)     & 81\% & 4.6X \\ 
& & \ourtool \xspace{} (with verifiable reward only)    & 87\% & 6.1X \\ 
& & \textbf{\ourtool \xspace{}}    & \textbf{94\%} & \textbf{9.02X} \\  
\midrule

\multirow{3}{*}{KernelBench}      
& \multirow{3}{*}{PyTorch $\rightarrow$ CUDA}
&   \ourtool \xspace{} (with structural reward only)     & 76\% & 3.7X \\ 
& & \ourtool \xspace{} (with verifiable reward only)    & 84\% & 4.3X \\ 
& & \textbf{\ourtool \xspace{}}    & \textbf{93\%} & \textbf{6.41X} \\  

\bottomrule
\end{tabular}%
}

\end{table}

\section{Related Works}

BabelTower \citep{wen2022babeltower} uses back-translation to improve C to CUDA translation using an unsupervised setting. CodeRosseta \citep{tehranijamsaz2024coderosetta} tackles the problem of C to CUDA translation by integrating AST Entity Recognition in the pre-training phase. QiMeng-MuPa \citep{ke2025qimeng} targets the same task of translating C to CUDA programs. They introduce a Mutual-Supervised Learning framework that works in an iterative manner, where a translation and a tester module generate data for each other to improve collectively. These works are not RL-based and do not include any reward signals. Also, these works mostly focus on C to CUDA kernel generation. But these approaches do not consider the problem of CUDA kernel optimization.

OptiML \citep{bhattacharjee2026optiml} proposes an inference-only framework to guide CUDA kernel optimization through the Monte Carlo Tree Search (MCTS) method; however, their approach is evaluated only on 5 tasks. Prior works have also demonstrated the effectiveness of using correctness and execution-based signals to guide learning in code generation and related domains \citep{chen2021evaluating, nijkamp2022codegen, cobbe2021training}. CodeV-R1  \citep{zhu2025qimeng} extends RLVR to structured domains by incorporating testbench-based verification and data synthesis pipelines to ensure correctness and scalability. However, these approaches primarily rely on outcome-based rewards like correctness or functional equivalence.

Reinforcement learning techniques like GRPO have been extensively applied to different domains like math \citep{wang2025reinforcement} and coding \citep{wei2025swe, waghjale2024ecco, shypula2023learning}. Also, the benefits of using multi-turn over single-turn RL is explored in various studies \citep{goldie2025synthetic, wang2025ragen, zhou2024archer, gehring2024rlef}.

Some recent works have explored RLVR for CUDA kernel generation and optimization. CUDA-L1 \citep{li2025cuda} introduces a contrastive RL framework that improves kernel performance using speedup-based rewards, demonstrating strong gains on KernelBench \citep{ouyang2025kernelbench} dataset. CUDA Agent \citep{dai2026cuda} proposes a large-scale agentic RL system that leverages automated profiling and verification environments to learn high-performance CUDA kernels. In parallel, multi-turn RL approaches such as \citep{baronio2025kevin} explicitly model iterative refinement by learning from sequences of execution feedback, highlighting the importance of trajectory-level optimization in CUDA kernels. KernelEvolve \citep{liao2025kernelevolve} is an automated, agent-driven system that generates and optimizes GPU kernels. It uses a search-based optimization process with execution feedback for generating optimized GPU programs. AKG Kernel Agent \citep{du2025akg} proposes a multi-agent collaborative framework that incorporates outcome-based signals like correctness and speedup to automate kernel generation and tuning. \citet{lange2025towards} proposed agentic workflows along wth LLM-based verification and efficient filtering for translating PyTorch programs to optimized CUDA kernels. Most prior work optimizes CUDA kernels using outcome-based signals such as correctness and speedup. However, these approaches rely purely on execution feedback and do not explicitly incorporate structural properties of parallel programs into the reward. In contrast, \ourtool \xspace{} introduces a composite reward that combines verifiable execution signals with structure-aware features derived from parallelization patterns.

\vspace{-10pt}

\section{Limitations}
One of the limitations of \ourtool \xspace{} is its training reliance on execution-based feedback, which introduces significant computational overhead. In particular, the need to repeatedly compile and benchmark candidate programs can be time-consuming, especially for large-scale kernels or diverse input configurations. As a result, this execution-driven reward formulation substantially increases the overall training time of the RL process. Another point worth mentioning is that \ourtool \xspace{} includes iterative refinement; it remains limited by the model and reward design, which can restrict its ability to discover optimization strategies beyond observed patterns.

\vspace{-10pt}

\section{Conclusion}

In this work, we introduced \ourtool, which leverages execution-based verifiable rewards with structure-aware signals derived from parallelization features to guide policy to generate optimized CUDA kernels. Extensive experiments across multiple benchmarks demonstrate that \ourtool \xspace{} consistently outperforms strong baselines in both correctness and performance. We believe \ourtool \xspace{} represents a step toward more principled and scalable learning frameworks for automated program optimization. Unlike purely black-box approaches, our method leverages fine-grained structural properties of programs, enabling the incorporation of interpretable, white-box reasoning signals derived from high-level performance characteristics signals derived from high-level over program performance.


\section*{Acknowledgment}

We would like to thank NSF for their generous support in funding this project (\#2426580) and (\#2422127). In addition, we extend our gratitude to Cisco AI Research for their generous support in the project. We thank the Research IT team of Iowa State University for providing access to HPC clusters for conducting the experiments of this research project.

{\small
\bibliographystyle{abbrvnat}
\bibliography{neurips_2026}
}

\medskip






\appendix

\section{Appendix}

\subsection{Structural Feature Extraction for Performance-Aware RL}
\label{app:feature-extraction}

To enable performance-aware reasoning without expensive execution, we design a structural feature extractor that maps a candidate CUDA program $y$ into a compact feature representation capturing GPU-relevant properties. These features are integrated into the reinforcement learning reward to guide the model toward efficient parallel implementations.

\subsection{Overview}

Given a CUDA program $y$, we extract a feature vector:
\begin{equation*}
\phi(y) = \big[
\text{coal}, \text{ai}, \text{occ}, \text{div}, \text{xfer}, 
\text{atomics}, \text{sync}, \text{kernels}, 
\text{tpb}, \text{gmem}, \text{ops}
\big] \in \mathbb{R}^{d}
\end{equation*}

Each component serves as a proxy for performance-critical aspects such as memory efficiency, compute intensity, and parallel scalability.

\subsection{Kernel Extraction}

We first identify CUDA kernels using pattern matching:
\begin{equation*}
\mathcal{K}(y) = \{(k_i, B_i)\}_{i=1}^{K}
\end{equation*}
where $k_i$ and $B_i$ denote the kernel name and body, respectively. The number of kernels is:
\begin{equation*}
\text{kernels}(y) = |\mathcal{K}(y)|
\end{equation*}

\subsection{Memory Coalescing}

For each global memory access $A[j]$, we analyze the index expression $j = f(\texttt{threadIdx.x}, \cdots)$ and approximate the stride:
\begin{equation*}
s = \frac{\partial j}{\partial \texttt{threadIdx.x}}
\end{equation*}

We define a coalescing score:
\begin{equation*}
\text{coal}(s) =
\begin{cases}
1.0 & s = 1 \\
0.4 & s = 2 \\
0.1 & s = 4 \\
-0.5 & s > 4 \\
0.2 & s = 0 \\
0.0 & \text{otherwise}
\end{cases}
\end{equation*}

Kernel-level and program-level scores:
\begin{equation*}
\text{coal}_k = \frac{1}{N_k} \sum_{i=1}^{N_k} \text{coal}(s_i),
\quad
\text{coal}(y) = \frac{1}{K} \sum_{k=1}^{K} \text{coal}_k
\end{equation*}

\subsection{Arithmetic Intensity}

We approximate arithmetic intensity as:
\begin{equation*}
\text{AI}_k = \frac{\text{ops}_k}{\text{gmem}_k + \epsilon}
\end{equation*}

To stabilize the scale:
\begin{equation*}
\text{ai}_k = \tanh(\lambda \cdot \text{AI}_k)
\end{equation*}
where $\lambda = 0.25$. The final value is averaged across kernels.

\subsection{Operation and Memory Counts}

We compute:
\begin{equation*}
\text{ops}_k = n_{+} + n_{-} + n_{\times} + n_{\div} + 4 \cdot n_{\text{popc}}
\end{equation*}

Global memory accesses are approximated as:
\begin{equation*}
\text{gmem}_k = \#\{\text{array accesses in } B_k\}
\end{equation*}

\subsection{Occupancy Proxy}

We approximate occupancy using threads-per-block (TPB):
\begin{equation*}
\text{occ}(y) =
\begin{cases}
1.0 & 128 \leq \text{TPB} \leq 256 \\
0.4 & 64 \leq \text{TPB} < 128 \ \text{or} \ 256 < \text{TPB} \leq 512 \\
-0.3 & \text{otherwise}
\end{cases}
\end{equation*}

\subsection{Control Divergence}

We penalize non-uniform branching:
\begin{equation*}
\text{div}_k = -\alpha \cdot N_{\text{non-uniform ifs}}
\end{equation*}
where conditionals dependent on thread or block indices are excluded.

\subsection{Synchronization and Atomics}

We directly count:
\begin{equation*}
\text{atomics}(y) = \sum_k n_{\text{atomic}}^{(k)}, \quad
\text{sync}(y) = \sum_k n_{\text{sync}}^{(k)}
\end{equation*}

\subsection{Data Transfer Overhead}

We penalize excessive host-device transfers:
\begin{equation*}
\text{xfer}(y) =
- \lambda_1 \max(0, N_{\text{memcpy}} - 2)
- \lambda_2 N_{\text{loop-memcpy}}
+ \lambda_3 \mathbb{I}_{\text{efficient}}
\end{equation*}

\subsection{Structural Reward}

We normalize features and compute a learned structural score:
\begin{equation*}
R_{\text{str}}(y) = w^\top \cdot \frac{\phi(y) - \mu}{\sigma}
\end{equation*}

This provides a dense, execution-free signal that correlates with performance.

\subsection{Feature Extraction Algorithm}

Algorithm \ref{abl:feat-extract-algo} shows the overall feature Extraction Algorithm.

\begin{algorithm}[t]
\caption{Structural Feature Extraction}
\label{abl:feat-extract-algo}
\begin{algorithmic}[1]
\STATE Input: CUDA program $y$
\STATE Extract kernels $\mathcal{K}(y)$
\FOR{each kernel $k \in \mathcal{K}(y)$}
    \STATE Count operations $\text{ops}_k$
    \STATE Count memory accesses $\text{gmem}_k$
    \STATE Compute arithmetic intensity $\text{ai}_k$
    \STATE Analyze memory stride $\rightarrow \text{coal}_k$
    \STATE Compute divergence $\text{div}_k$
    \STATE Count atomics and synchronization
\ENDFOR
\STATE Estimate occupancy from TPB
\STATE Compute transfer cost $\text{xfer}(y)$
\STATE Aggregate features into $\phi(y)$
\RETURN $\phi(y)$
\end{algorithmic}
\end{algorithm}

\subsection{MLP Ranker Architecture}
\label{app:ranker-arch}
\begin{figure}
    \centering
    \includegraphics[width=\linewidth]{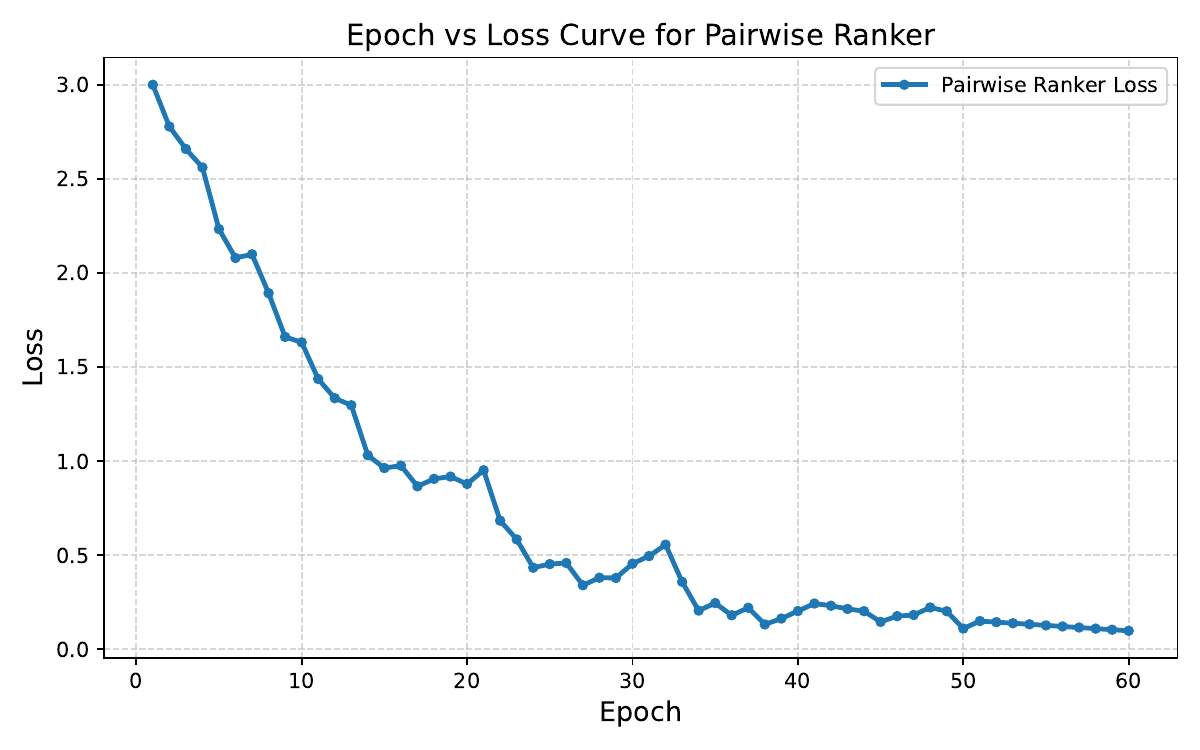}
    \vspace{-20pt}
    \caption{Epoch vs loss curve for Pairwise Ranking MLP model}
    \label{figs:app-epoch vs loss}
\end{figure}

We employ a lightweight multi-layer perceptron (MLP) ranker to learn a scalar quality score for each CUDA program candidate. The network consists of fully connected layers with ReLU activations and dropout regularization. Given a normalized feature vector \(\phi(y)\), the model predicts a ranking score:

\begin{equation*}
s(y) = f_{\theta}(\phi(y)),
\end{equation*}

where \(f_{\theta}\) denotes the parameterized MLP ranker.

The default architecture uses hidden dimensions of \(64\) and \(32\) with a dropout probability of \(0.1\). The model is optimized using AdamW with learning rate \(3 \times 10^{-4}\), weight decay \(10^{-4}\), batch size \(512\), and gradient clipping threshold \(1.0\). It can be observed from Figure \ref{figs:app-epoch vs loss} that training loss saturates after 50 epochs.

\subsection{Training Times}

The MLP based offline ranker needs 2 hours 16 minutes to train. C-to-CUDA RL training requires 3 days 12 hours and 12 minutes. PyTorch-to-CUDA training requires 3 days 8 hours and 44 minutes. All models are trained on two A100 GPUs with 40 GB GPU memory.

\subsection{Baseline Sources}

For CodeLlama-34b-Instruct-hf (https://huggingface.co/meta-llama/CodeLlama-34b-Instruct-hf), Qwen2.5-Coder (https://huggingface.co/Qwen/Qwen2.5-Coder-32B), Qwen-3-32B (https://huggingface.co/Qwen/Qwen3-32B) models are accessed using the HuggingFace API. We use the $agent\_workdir$ for CUDA Agent execution from their open-source GitHub repository (https://github.com/BytedTsinghua-SIA/CUDA-Agent/tree/main). For Kevin we also use the HuggingFace API (https://huggingface.co/cognition-ai/Kevin-32B). For QiMeng-MuPa we use the translator-Llama-3-8B model (https://huggingface.co/kcxain/translator-Llama-3-8B) which is a finetuned version of the Meta-Llama-3-8B model. We choose this model as this version produces the best results on the BabelTower dataset.

\subsection{Results}
\label{app:results-pass}.

Here we provide more detailed results for the datasets and the baseline models using Pass@{1, 3, 5} metrics. It can be observed from Table \ref{tbl:app-pass-results-main} that \ourtool \xspace{} achieves better results in most cases outperforming strong baselines like Qwen-3-32B \citep{qwen3technicalreport} model, Cuda Agent \citep{dai2026cuda} and QiMeng-MuPa \citep{ke2025qimeng} across different dataset.

\begin{table}[]
\centering
\label{tbl:app-pass-results-main}
\caption{Comparing performance of \ourtool \xspace{} across 3 different datasets using Pass@{1, 3, 5} metrics.}
\resizebox{\textwidth}{!}{%
\begin{tabular}{l l c c c c c c}
\hline
\multirow{2}{*}{Dataset} & \multirow{2}{*}{Model} & \multicolumn{3}{c}{\begin{tabular}[c]{@{}c@{}}Correctness \\ (All test cases passed)\end{tabular}} & \multicolumn{3}{c}{\begin{tabular}[c]{@{}c@{}}Performance \\ (Speedup)\end{tabular}} \\ 
\cline{3-8}
 &  & Pass@1 & Pass@3 & Pass@5 & Pass@1 & Pass@3 & Pass@5 \\ 
\hline

\multirow{6}{*}{CUDA-Perf-Dataset} 
& OpenAI O4-mini \citep{openai_api}     & 31\% & 36\% & 41\% & 0.39X & 0.95X & 1.3X \\ 
& CodeRosetta \citep{tehranijamsaz2024coderosetta} & 29\% & 31\% & 45\% & 0.48X & 0.96X  & 1.5X \\ 
& CodeLlama-34b-Instruct-hf \citep{roziere2023code} & 29\% & 37\% & 53\% & 0.58X & 1.5X  & 3.5X \\ 
& Qwen2.5-Coder-32B \citep{hui2024qwen2}    & 23\% & 42\% & 61\% & 0.81X & 1.6X  & 5.3X \\ 
& QiMeng-MuPa \citep{ke2025qimeng}     & 20\% & 49\% & 63\% & 0.91X & 2.1X  & 5.9X \\ 
& Qwen-3-32B \citep{qwen3technicalreport}      & 31\% & 44\% & 72\% & 1.2X  & 2.3X  & 6.2X \\ 
& \textbf{\ourtool}        & \textbf{45\%} & \textbf{64\%} & \textbf{89\%} & \textbf{3.1X } & \textbf{5.7X } & \textbf{11.02X} \\ 
\hline

\multirow{6}{*}{BabelTower} 
& OpenAI O4-mini \citep{openai_api}    & 29\% & 31\% & 52\% & 0.88X & 0.91X & 1.9X \\ 
& CodeLlama-34b-Instruct-hf \citep{roziere2023code} & 28\% & 44\% & 68\% & 0.77X & 1.2X  & 3.1X \\ 
& CodeRosetta \citep{tehranijamsaz2024coderosetta} & 40\% & 57\% & 71\% & 0.79X & 1.9X  & 3.2X \\ 
& Qwen2.5-Coder-32B \citep{hui2024qwen2}   & 56\% & 67\% & 79\% & 0.97X & 1.3X  & 4.4X \\ 
& Qwen-3-32B \citep{qwen3technicalreport}      & 66\% & 72\% & 86\% & 1.1X  & 1.9X  & 4.92X \\ 
& QiMeng-MuPa \citep{ke2025qimeng}     & \textbf{82\%} & 89\% & \textbf{94\%} & 1.2X  & 2.4X  & 6.01X \\ 
& \textbf{\ourtool}        & 76\% & \textbf{91\%} & \textbf{94\%} & \textbf{2.5X}  & \textbf{4.9X}  & \textbf{9.02X} \\ 
\hline

\multirow{7}{*}{KernelBench} 
& OpenAI O4-mini \citep{openai_api}    & 26\% & 33\% & 38\% & 0.27X & 0.44X & 0.78X \\ 
& CodeLlama-34b-Instruct-hf \citep{roziere2023code} & 22\% & 27\% & 41\% & 0.17X & 0.22X & 0.25X \\ 
& Qwen2.5-Coder-32B \citep{hui2024qwen2}    & 31\% & 38\% & 51\% & 0.21X & 0.26X & 0.5X \\ 
& Qwen-3-32B \citep{qwen3technicalreport}     & 29\% & 41\% & 56\% & 0.22X & 0.25X & 0.53X \\ 
& Kevin   \citep{baronio2025kevin}        & 39\% & 71\% & 82\% & 0.68X & 0.99X & 1.10X \\ 
& Cuda Agent \citep{dai2026cuda}     & 34\% & 69\% & 86\% & 1.2X  & 2.3X  & 3.09X \\ 
& \textbf{\ourtool}       & \textbf{55}\% & \textbf{82}\% & \textbf{93}\% & \textbf{1.9X}  & \textbf{2.8X}  & \textbf{6.41X} \\ 
\hline

\end{tabular}%
}
\end{table}

\newpage

\subsection{Prompts for Policy Model}
\label{app:prompt}

Below is the skeleton for the prompts that are used in the RL training phase for policy models to generate CUDA programs.

\begin{large}
\begin{verbatim}
[FEWSHOT]
[EXAMPLE]
[SEQ_CODE]
<Sequential C/PyTorch program>
[/SEQ_CODE]

[CUDA]
<Optimized CUDA implementation>
[/CUDA]

[INPUTS]
<Test inputs>
[/INPUTS]
[SPEEDUP]
<reported speedup>
[/SPEEDUP]
[/EXAMPLE]

[EXAMPLE]
[SEQ_CODE]
<Sequential C/PyTorch program>
[/SEQ_CODE]

[CUDA]
<Optimized CUDA implementation>
[/CUDA]

[INPUTS]
<Test inputs>
[/INPUTS]
[SPEEDUP]
<reported speedup>
[/SPEEDUP]
[/EXAMPLE]
...
[/FEWSHOT]

[FEEDBACK_HISTORY]
<Summarized feedback of previous turns>
[/FEEDBACK_HISTORY]

[FEEDBACK]
<Most recent feedback>
[COMPILE_ERROR]
[/COMPILE_ERROR]
[FAILED_CASE]
[/FAILED_CASE]
[/FEEDBACK]
Convert the following sequential program into an efficient
CUDA C++ program that produces the same output on
the given hardware.

When generating the program try to use 
different optimization strategies including: 
memory coalescing,
arithmetic intensity & operation count, 
occupancy (threads per block), 
control divergence, synchronization, 
atomic operations and data transfer overhead. 

Use the codes within <EXAMPLE></EXAMPLE> block as some references.

Use the information in [FEEDBACK][/FEEDBACK] 
block to improve the CUDA program.

Only output CUDA code compilable by nvcc as a
single translation unit. 

Output CUDA code within a <OUTPUT_CUDA></OUTPUT_CUDA> Block

[SEQ_CODE]
<Target sequential program>
[/SEQ_CODE]

[INPUTS]
<Test inputs>
[/INPUTS]

\end{verbatim}
\end{large}


\end{document}